%% file: main.tex
\pdfoutput=1

\documentclass[11pt]{article}

\usepackage[]{acl}
\usepackage{xcolor}
\usepackage{graphicx}
\usepackage{times}
\usepackage{latexsym}
\usepackage{amsthm}
\newtheorem*{remark}{Remark}
\usepackage[T1]{fontenc}

\usepackage[utf8]{inputenc}

\usepackage{microtype}

\usepackage{inconsolata}

\include{Preamble/custom_packages}
\include{Preamble/custom_commands}

\title{\ourmodel{}: A pioneering Large Language Model for Law} 

\author{Pierre Colombo$^{1,2,*}$ \quad Telmo Pessoa Pires$^{1,*}$ \quad Malik Boudiaf$^{1,*}$  \\
\textbf{Dominic Culver}$^{1,*}$  \quad   \textbf{Rui Melo}$^{1,*}$ \quad 
\textbf{Caio Corro}$^3$   \quad  \textbf{André F. T. Martins}$^4$  \\ 
\textbf{Fabrizio Esposito}$^5$ \quad \textbf{Vera Lúcia Raposo}$^5$  \quad
\textbf{Sofia Morgado}$^1$ \quad
\textbf{Michael Desa}$^1$ \\
$^1$Equall.ai, New York, Paris, Lisbon \\
$^2$MICS, CentraleSupélec, Université Paris-Saclay\\
$^3$Sorbonne Université, CNRS, ISIR, Paris\\
$^4$Instituto Superior Técnico, Universidade de Lisboa\\
$^5$ NOVA School of Law, Lisboa \\
\texttt{firstname@equall.ai}
}

\begin{document}
\maketitle
\def\thefootnote{*}\footnotetext{Equal contribution.}\def\thefootnote{\arabic{footnote}}
\begin{abstract}
In this paper, we introduce \ourmodel{}, a large language model (LLM) tailored for the legal domain. With 7 billion parameters, \ourmodel{} is the first LLM designed explicitly for legal text comprehension and generation. Leveraging the Mistral 7B architecture as its foundation, \ourmodel{} is trained on an English legal corpus of over 30 billion tokens. \ourmodel{} exhibits state-of-the-art proficiency in understanding and processing legal documents. Additionally, we present a novel instructional fine-tuning method that leverages legal datasets to further enhance \ourmodel{}'s performance in legal tasks. \ourmodel{} is released under the MIT License.
\end{abstract}

\section{Introduction}

In the rapidly evolving landscape of artificial intelligence, the applications of large language models (LLMs) \cite{achiam2023gpt,scao2022bloom,penedo2023refinedweb,touvron2023llama,jiang2023mistral,jiang2024mixtral,touvron2023llama2,bai2023qwen} have witnessed large advancements across various domains, like \textit{e.g.}\ translation \cite{xu2023paradigm}, medical \cite{chen2023meditron}, and code generation \cite{roziere2023code,li2023starcoder}. From natural language processing to machine translation, these models have exhibited exceptional capabilities in understanding and generating human-like text \cite{weber2023testing,islam2023distinguishing,mitchell2023detectgpt}. 
However, one field that has yet to experience the full benefit of this transformative technology is the legal domain \cite{martin2024better,licari2022italian}. As legal professionals grapple with an ever-expanding volume of complex documents, there is a growing need for a dedicated LLM that can help navigate and interpret legal material \cite{savelka2023explaining,katz2023gpt,xiao2021lawformer}.

In this paper, we present a pioneering initiative to develop the first legal LLM publicly available.
Legal text, characterized by its unique syntax and specialized vocabulary presents a distinct linguistic challenge \cite{chalkidis2020legal,niklaus2021swiss}.
Our approach focuses on extensive pretraining \cite{gururangan2020don,yao2021adapt} using dedicated legal corpora from English-speaking jurisdictions such as the USA, Canada, the UK, and Europe \cite{aletras2016predicting,gutierrez2021spanish}.
Leveraging the pretraining on a large and diverse legal dataset, both scraped by our team as well as from previous literature \citep{niklaus2022budgetlongformer}, our LLM, \ourmodel{}, aims not only to comprehend the complexities of legal documents but also to adapt to the evolving nature of legal discourse.


By focusing on the needs of legal practitioners and harnessing the power of pretraining on dedicated legal corpora, our work represents an important step towards fulfilling the unique demands of the legal domain. We anticipate that introducing the first LLM for law will not only empower legal professionals but also catalyze further innovation at the intersection of artificial intelligence and the legal community - making a significant contribution to legal language understanding and application \cite{prakken2013logical}. We summarize the contributions of this work as follows:

\paragraph{Contribution 1: A family of legal LLMs.} In this paper, we introduce the \ourmodel{}'s family, a collection of Legal Language Models meticulously crafted to tackle the distinctive challenges encountered within the legal domain. We unveil \ourmodel{}, a 7-billion-parameter language model specifically tailored to legal text. With its specialized training regimen, \ourmodel{} demonstrates a superior understanding of the nuances in legal language compared to generic models. Furthermore, we release \ourmodelift{}, an instruction-tuned variant, carefully engineered to outperform existing models such as \texttt{Mistral} or \texttt{Llama} on a variety of legal tasks\footnote{Model is available at \url{https://huggingface.co/Equall}.}.

\paragraph{Contribution 2: An improved evaluation protocol for legal LLMs.} Concurrently, we introduce \legalbench{}, a supplemental iteration of LegalBench \cite{guha2022legalbench,guha2023legalbench}\footnote{Dataset is processed and available at \url{https://huggingface.co/Equall}}, crafted to better gauge and refine the legal proficiency of language models, which we hope will contribute to future advancements into research in the legal domain. To further enrich the models' capabilities in legal contexts, we also include the legal tasks of the popular MMLU benchmark \cite{hendrycks2020measuring} in our evaluation protocol, particularly focusing on international law, professional law\footnote{We use the term ``professional law'' here as defined in \citep{hendrycks2020measuring}} and jurisprudence.
\begin{figure*}[!ht]
    \centering
    \includegraphics[width=\textwidth]{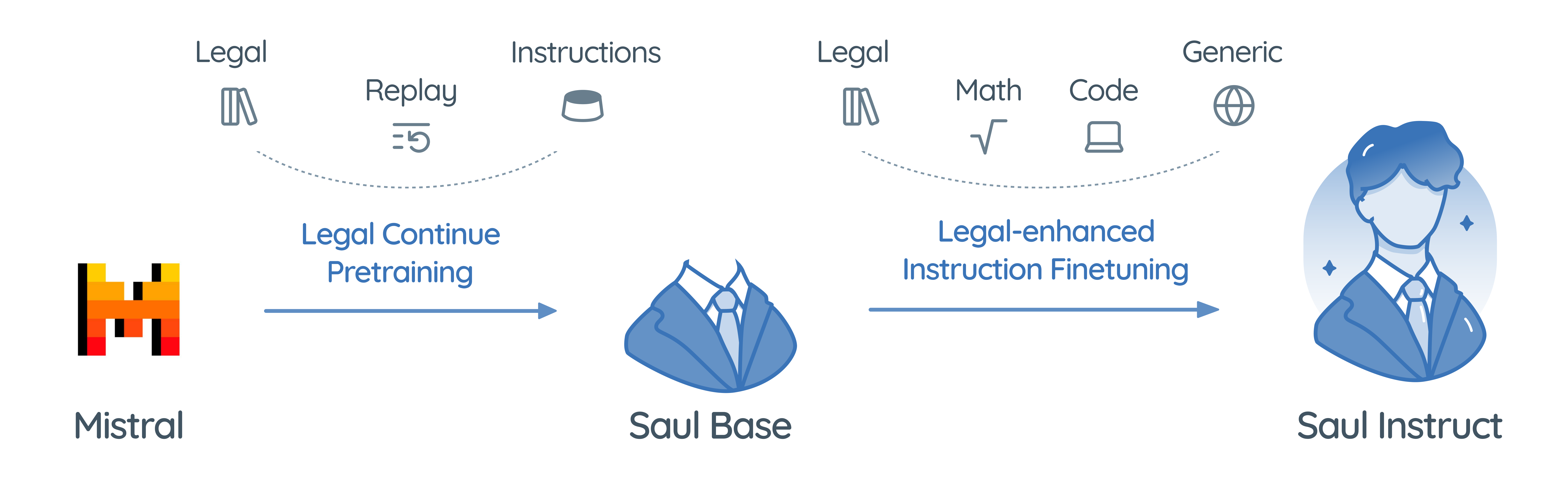}
    \caption{\textbf{Procedure for constructing \ourmodel{}}. We rely on legal datasets augmented with replay data, and instructions datasets. For fine-tuning we enrich our instruction finetuning dataset further with legal instructions.}
    \label{fig:main_saul}
\end{figure*}

\paragraph{Contribution 3: Model, Evaluation Code \& Licensing.} To foster widespread adoption and promote innovation, we release \ourmodel{} and \ourmodelift{}, as well as our evaluation code under the MIT License. This open licensing approach encourages collaborative development and adoption into a wide array of commercial and research endeavors within the legal domain and beyond.

\section{\ourmodel{}: Extending the legal capabilities of Language Models}


A wide range of open-source large language models is available for the backbone, spanning from $70$ million parameter models like Pythia \citep{biderman2023pythia} to $180$ billion parameter models like Falcon \citep{falcon}. In this work, we choose the Mistral $7$B model, a $7$ billion parameter open-source model that achieves high performance across benchmarks and tasks \citep{jiang2023mistral}.

Our methodology, shown in \autoref{fig:main_saul} involves a two-step process that we describe below.

\subsection{Enhancing Mistral's Legal Capabilities}
While generic models \cite{touvron2023llama,taylor2022galactica,zhang2022opt,gu2023mamba,falcon,zhang2024tinyllama,faysse2024croissantllm} gain some exposure to legal data during their training, it typically only represents a minor fraction of the overall data. A straightforward method to enhance performance for legal tasks is to perform additional training focusing on legal data. This approach, particularly focused on decoder models, has been successfully used in various fields such as medicine \cite{chen2023meditron,ji2023domain}, translation \cite{xu2023paradigm,wu2024adapting}, and coding \cite{roziere2023code}. 
The key advantage of this approach is its scalability and independence from the specific characteristics of the training data.
Other research on domain adaptation has attempted to specialize language models via pretext tasks. However, these efforts often rely on smaller-scale approaches \cite{niklaus2023can}, are computationally expensive \cite{vu2020effective,lu2023prompt}, or lack scalability \cite{cheng2023adapting,cui2023chatlaw,nishida2019unsupervised}.

For these reasons, as well as the availability of large-scale legal corpora from the web, we chose to focus on \emph{continued pretraining}.
We meticulously curate a high-quality dataset sourced from diverse legal content repositories. After rigorous filtering 
 \citep{penedo2023refinedweb} and deduplication \citep{chenghao_mou_2023_8364980,kocetkov2023the}, we end up with a corpus of $30$ billion tokens, which serves as a robust foundation for continued pretraining.
 
\subsection{Improving Legal Instruction Following}
To support user requests and conversational interaction, LLMs typically undergo instruction tuning, a critical process involving training on supervised conversational pairs. This step is essential for crafting a versatile model, adept at addressing user queries \cite{wang2023far,wei2021finetuned,chung2022scaling,Faysse_2023,ding2023enhancing,wang2023selfinstruct}.

For general-purpose language models, diversity and quality of instruction are crucial \cite{cao2023instruction,zhou2023lima}. However, in specialized domains it is crucial to incorporate task-specific and specialized prompts to enhance performance. Our instruction fine-tuning stage involves $2$ key components: generic (ie, non-legal) and legal instructions. The former help enhance the model's understanding and following of commands, and includes data from diverse domains such as coding, mathematics, and general conversations. For the latter we employ an extensive collection of datasets tailored to the nuances of legal domains, covering legal question answering and summarization, among others.
Through this meticulous fine-tuning on instructional data, our model, \ourmodelift{}, is able to grasp legal intricacies and excels in a wide range of associated tasks.


\begin{remark}
It's worth noting that many common LLMs \cite{tunstall2023zephyr} include an additional step of to align the model with human preference \cite{rafailov2023direct,munos2023nash,vonwerra2022trl}. In our case, early experiments did not show any meaningful improvement in performance and so we opted to not pursue this avenue for the present paper.
\end{remark}

\section{Data}
In this section we describe our data collection and cleaning schemes.

\subsection{Legal Pretraining Corpora}
Unlike fields such as science and medicine, the legal landscape varies significantly across countries and jurisdictions, reflecting differences not only in local laws but also in legal traditions, like common law versus civil law \cite{henderson2022pile}. Thus, we gathered legal texts from various jurisdictions, with a primary focus on the English language due to its widespread use in legal contexts worldwide. Our collection includes data from the U.S. \cite{tuggener2020ledgar}, Europe \cite{chalkidis2019neural}, and Australia \cite{butler-2023-open-australian-legal-corpus}, covering a diverse range of legal systems. Through this thorough curation process and aggressive cleaning (see \Cref{sec:data_cleaning}), we end up with a corpus of 30 billion tokens, capturing the intricacies of legal language across regions.

\subsubsection{Dataset Composition} \label{sec:dataset_composition}

\paragraph{Legal Sources}
We combine both previously available datasets, such as the FreeLaw subset from The Pile \citep{gao2020pile} and MultiLegal Pile \citep{niklaus2023multilegalpile}, as well as data scraped from publicly available sources on the Web. We list the different sources of data in \Cref{tab:data-sources}.

\begin{table}[ht]
    \small
    \centering
    \begin{tabular}{lc}
    \toprule              
     Name & Tokens \\
     \midrule
     FreeLaw\tablefootnote{We used the subset from The Pile \citep{gao2020pile}.} & $15$B \\
     EDGAR\tablefootnote{\url{https://www.sec.gov/edgar}} & 5B \\
     English MultiLegal Pile\tablefootnote{We limited ourselves to the commercially-licensed subset: \url{https://huggingface.co/datasets/joelniklaus/Multi_Legal_Pile_Commercial}} & $50$B \\
     English EuroParl \citep{koehn-2005-europarl} & $6$B \\
     GovInfo\tablefootnote{\url{https://www.govinfo.gov/}} Statutes, Opinions \& Codes & $11$B \\
     Law Stack Exchange\tablefootnote{\url{https://huggingface.co/datasets/ymoslem/Law-StackExchange}} & $19$M \\
     Commercial Open Australian Legal Corpus\tablefootnote{\url{https://github.com/umarbutler/open-australian-legal-corpus-creator}} & $0.5$B \\
     EU Legislation\tablefootnote{Scraped from \url{https://eur-lex.europa.eu/homepage.html}} & $315$M \\
     UK Legislation\tablefootnote{\url{https://www.legislation.gov.uk/}} & $190$M \\
     Court Transcripts\tablefootnote{Obtained from CourtListener: \url{https://www.courtlistener.com/}. We use Whisper \citep{radford2022robust} to transcribe the audio files.} & $350$M \\
     UPSTO\tablefootnote{\url{https://bulkdata.uspto.gov/}} & $4.7$B \\
     Total & $94$B \\
    \bottomrule
    \end{tabular}
    \caption{\textbf{Sources of Legal Pretraining Data.} These sources contain noise and heavily duplicated documents, which we filtered and deduplicated, resulting in a 30 billion tokens dataset.}
    \label{tab:data-sources}
\end{table}

There is quite a lot of overlap between the different sources, and we run very aggressive cleaning and deduplication steps, described in \Cref{sec:data_cleaning}.

\paragraph{Replay Sources}
To reduce the risk of catastrophic forgetting \citep{MCCLOSKEY1989109} during continued pretraining, we incorporate data from the prior training distribution, following prior literature \cite{chen2023meditron,sun2020distill}. However, since the training data for Mistral is undisclosed, we introduce commonly available ``general'' data from Wikipedia, StackExchange, and GitHub, comprising roughly $2\%$ of the final training mix. These datasets are sampled from SlimPajama \cite{shen2023slimpajama,together2023redpajama,soboleva2023slimpajama}.

\paragraph{Instruction Sources}
Additionally, we found it beneficial to include conversational data during pretraining. This is inspired by recent advances in neural machine translation, which highlight that the robust capabilities of LLMs in translation are due to the existence of accidental parallel data in the training corpus \cite{anil2023palm,briakou2023searching}. Specifically, this means that we include the Super Natural Instruction \cite{wang2022super} and FLAN collection \cite{longpre2023flan} during pretraining.


\subsubsection{Data Cleaning}
\label{sec:data_cleaning}

A significant fraction of the collected data is either in PDF files or is text extracted from PDFs\footnote{We used \href{https://poppler.freedesktop.org/}{Poppler} for text extraction from PDF files.}. This means that the text has some artifacts, including i) page numbers in the middle of sentences; ii) line numbers; iii) non-normalized unicode characters; iv) broken lines of text; v) repeated characters: new lines, dashes, etc; vi) other artifacts. We addressed these issues using a combination of rules and heuristics to filter the data.

\paragraph{Text Normalization}
We normalize all unicode with the NFKC method, available through the \texttt{unicodedata} Python package.

\paragraph{Rule filters}
Following \citet{elazar2023whats}, we found the most common 10-grams in our dataset and used regular expressions to remove the undesired ones, which were mostly repeated characters. Concretely, $8$ of the top $10$ 10-grams in the original data were repeated characters, eg: ``\texttt{- - - - - - - - - -}'', ``\texttt{. . . . . . . . . .}'', or ``\texttt{* * * * * * * * * *}'', and weird characters, ie encoding issues. Additionally, we removed repeated whitespace (spaces, new lines, and tabs), as well as any HTML tag that made it through our pipeline.

\paragraph{Perplexity filtering}
We trained a KenLM model \citep{heafield-2011-kenlm} on a small subset of carefully inspected legal data, and used it to filter any high perplexity paragraph. This removed non-English text as well as most of the ``weird'' unicode sequences present in the data. We show some of the most common $10$-grams in the filtered data on \Cref{tab:common-ngrams}.

\begin{table}[ht]
    \scriptsize
    \centering
    \begin{tabular}{c}
    \toprule              
    Common 10-grams \\
    \midrule
    \texttt{have been obvious to one of ordinary skill in the} \\
    \texttt{before the effective filing date of the claimed invention to} \\
    \texttt{rejected under 35 U.S.C . 103 as being unpatentable over} \\
    \bottomrule
    \end{tabular}
    \caption{\textbf{Most common 10-grams} in the pretraining dataset.}
    \label{tab:common-ngrams}
\end{table}

\subsubsection{Data Deduplication}
Inspired by \citet{kocetkov2023the,lee2021deduplicating}, we removed duplicates and near-duplicates from the training data using \citet{chenghao_mou_2023_8364980}, with default parameters, after which we were left with roughly $30$B tokens of high-quality text.

\subsection{Instruction Finetuning Mixes}
Instruction fine-tuning is crucial for getting the best performance out of the pre-trained decoder models across different tasks. We use a mix of general and legal instructions to train the model to understand and follow instructions well, with a focus on legal expertise. 

\paragraph{General Instructions}
When it comes to general instructions, we gather them from four primary sources:
\begin{enumerate}
    \item \textbf{SlimOrca} This subset of the FLAN collection comprises generic instructions, offering a focused resource for various tasks \cite{mukherjee2023orca,SlimOrca}.
    \item \textbf{Meta Math Question Answering Instructions} Designed for mathematical inquiry, this dataset\footnote{Accessible at \url{meta-math/MetaMathQA}} presents a range of mathematical questions, facilitating research in math-based natural language processing \cite{yu2023metamath}.
    \item \textbf{General Conversations from UltraChat} Capturing diverse conversational contexts, this GPT-derived dataset contributes to enhancing natural language understanding and generation systems \cite{ding2023enhancing}.
    \item \textbf{Code Instructions from Glaive Code Assistant v2\footnote{Available at \url{https://huggingface.co/datasets/glaiveai/glaive-code-assistant-v2}}} Training on code has been shown to increase the reasoning ability of models \citep{ma2023training}
\end{enumerate}

We meticulously filter, deduplicate, and curate all this data, resulting in a refined dataset comprising $600$K instructions.
 
\begin{figure}
    \centering
    \includegraphics[width=\linewidth]{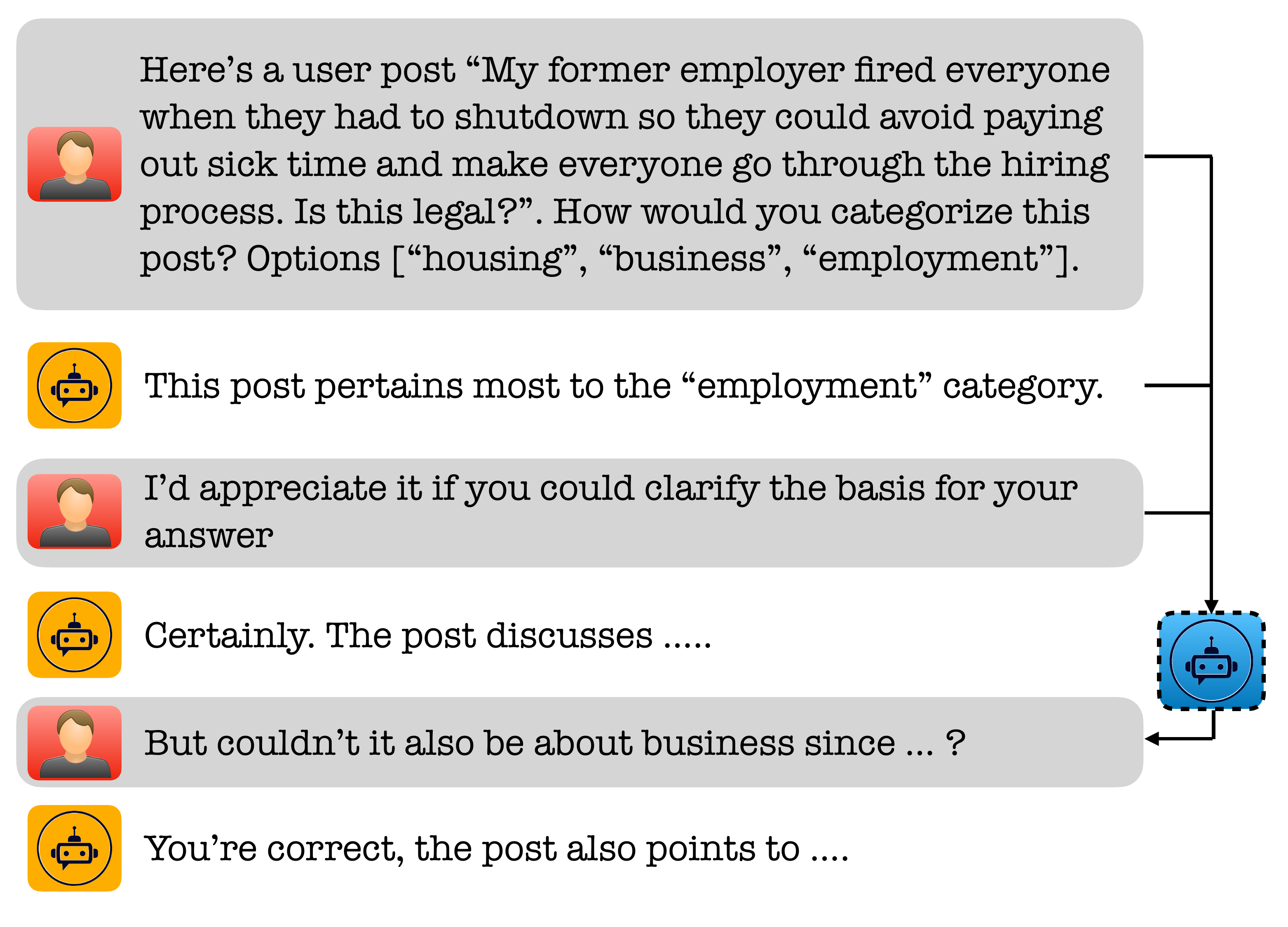}
    \caption{\textbf{Turning dataset with metadata into a conversation.} Taking the example of Reddit post classification, we turn a labeled example \{"\textit{My employer fired me because \dots Is it legal?}", "\textit{employment}" \}, we hard-code the first three turns of the conversation by simply reformulating the query and answer as a natural conversation. We then complete the conversation using a \textit{user} model(blue dashed), whose task is to continue generating relevant questions from the ongoing conversation, and an \textit{assistant} model that provides answers. Both \textit{assistant} and \textit{user} models are \texttt{Mistral-7B-instruct}.}
    \label{fig:ift_generation}
\end{figure}

\paragraph{Legal Instruction Construction}
We synthetically generate comprehensive conversations addressing fundamental legal competencies across multiple legal document types \cite{ding2023enhancing}. We leverage a \texttt{Mistral-7B-instruct} to transform legal texts augmented with metadata into coherent conversations.
The methodology involves initiating the conversation with $3$ predefined turns: (1) the user articulates a request related to the legal document, (2) the assistant responds by rephrasing the metadata (e.g., document type, date, name of a judge), and (3) the user prompts the assistant to elaborate on its reasoning. Subsequently, we extend the conversation through a series of turns, where a user model progressively poses more specific questions to grasp the assistant's reasoning. Simultaneously, an assistant model provides in-depth insights. An illustrative example is presented in \Cref{fig:ift_generation}. Notably, we ensure the exclusion of the test set from existing benchmarks.


\section{Evaluation of Legal Knowledge}
\label{sec:legal_bench_instruct}

To evaluate the model's legal abilities, we use $3$ benchmarks (i) we compare the perplexity of the backbones on $5$ types of legal documents, (ii) we enhance LegalBench with \legalbench{} for deeper evaluation, (iii) we rely on the legal section of MMLU for additional insights.

\paragraph{Perplexity Measurement}
To evaluate the adaptability of the backbones to legal documents, we assess perplexity using benchmark datasets spanning four distinct legal domains: \emph{contracts, judicial decisions, opinion text, and legislation}. We ensure that the datasets are up-to-date, and sourced after the collection cut-off date from LLM data. Specifically, contract data is sourced from EDGAR (first quarter of 2024), legal decisions from ICSID court decisions published after October 2023, legislation focuses on US bills submitted before the House or Senate after October 2023, and party submissions include Texas briefs submitted after October 2023.


During our investigations, we found a significant limitation in the original prompts of LegalBench. The complex nature of these prompts, combined with the challenges encountered by open source LLMs in adhering to instructions - particularly in handling formatting - leads to a substantial drop in performance (as measured by accuracy). The generated sentences are often verbose and difficult to parse, rendering LegalBench in its current form too stringent and failing to accurately gauge improvement on the task. 

For example, in some of the tasks, performance is evaluated by the first word the model predicts, and this word is expected to be a \emph{Yes/No}. This means that if the response is a bit verbose it will be counted as incorrect, even if a human would classify it as a correct answer. To remedy this shortcoming, we refine the prompts by 1) removing distracting few-shot examples and 2) concluding with a specific instruction for the model to generate tags (see \Cref{tab:legalbench}).

\begin{table}[ht]
    \scriptsize
    \centering
    \begin{tabular}{p{7cm}}
    \toprule              
    \textbf{Original Prompt} \\
    \midrule
    The Telemarketing Sales Rule is provided by 16 C.F.R. § 310.3(a)(1) and 16 C.F.R. § 310.3(a)(2).
    
\\\textbf{Question:} Acme Toys is a telemarketer subject to the Telemarketing Sales Rule. Acme Toys told a customer that its frisbees cost \$10 each, when in fact the frisbees cost \$12 each. The customer agreed to the sale and was charged \$12. Is this a violation of the Telemarketing Sales Rule?
\\\textbf{Answer:} Yes

\\\textbf{Question:} Acme Toys is a telemarketer subject to the Telemarketing Sales Rule. Acme Toys told a customer that its frisbees cost \$10 each, when in fact the frisbees did cost \$10, but Acme Toys did not disclose that shipping would cost an additional \$5. The customer agreed to the sale. Is this a violation of the Telemarketing Sales Rule?
\\\textbf{Answer:} Yes

\\\textbf{Question:} Acme Industrial Products is a telemarketer subject to the Telemarketing Sales Rule. Acme Industrial Products told a customer that its brooms cost \$12 each, and the brooms did in fact cost \$12. The customer agreed to the sale. Is this a violation of the Telemarketing Sales Rule?
\\\textbf{Answer:} No

\\\textbf{Question:} Acme Industrial Products is a telemarketer subject to the Telemarketing Sales Rule. Acme Industrial Products told a customer that it would sell them 4 brooms for \$10 and that shipping would be \$5. Then, the customer agreed to the sale. Is this a violation of the Telemarketing Sales Rule?
\\\textbf{Answer:} No

\\\textbf{Question:}  \{{text\}}
\\\textbf{Answer:}\\
\\\hline           
\textbf{Curated Prompt} (Ours) \\\hline 

The Telemarketing Sales Rule is provided by 16 C.F.R. § 310.3(a)(1) and 16 C.F.R. § 310.3(a)(2).

Answer the following question: \{text\}

\textit{Answer by only outputting "Yes" or "No"} \\
\bottomrule
\end{tabular}
    \caption{\textbf{Example from \legalbench{}}. We manually curated and corrected typos, removing a few short examples from LegalBench as they were found to distract LLMs of size 7B.\vspace{-.5cm}}
    \label{tab:legalbench}
\end{table} 

\paragraph{Massive Multitask Language Understanding (MMLU)} The MMLU benchmark \citep{hendrycks2020measuring} has been widely employed to gauge the advances in LLM performance. In our study, we center our analysis on the legal domain, with a specific focus on:
\textit{international law}, \textit{professional law}, and \textit{jurisprudence}. Those tasks respectively contain $120$, $1500$, and $110$ examples.

\subsection{Metrics}
We use the same metric as the original LegalBench \cite{guha2023legalbench} paper: balanced accuracy. Balanced accuracy allows for handling better-imbalanced classification tasks, such as the ones presented in both benchmarks. We also use balanced accuracy for the legal tasks of MMLU. Unless otherwise noted, any score reported throughout this section refers to the balanced accuracy.

\section{Experimental Setting}
\begin{figure}
    \centering
    \includegraphics[width=0.9\linewidth]{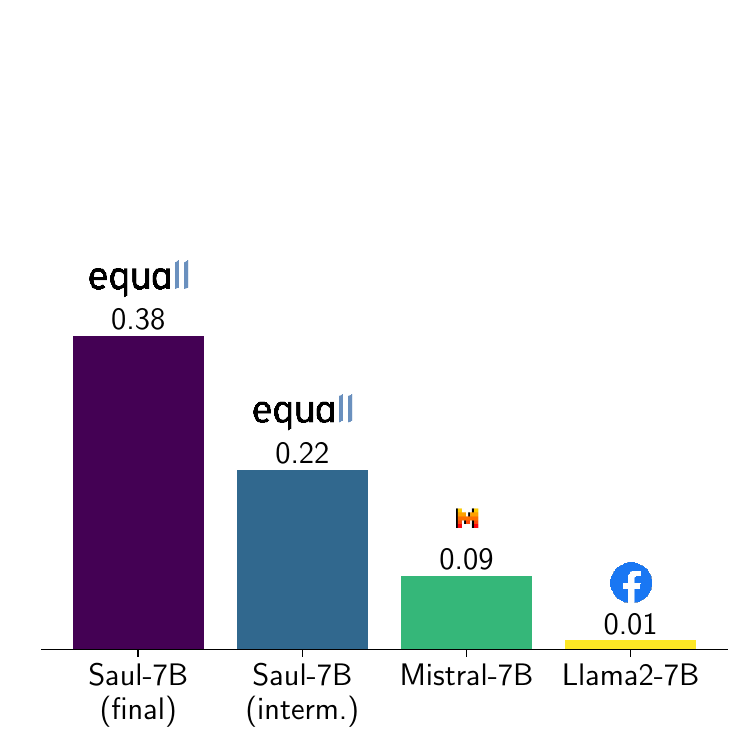}
    \caption{\textbf{Performance of base models on \legalbench{}.} Interestingly, although not instruction fine-tuned, \ourmodel{} is still able to achieve impressive improvements on the benchmark, compared to other base models, including \ourmodel's initial checkpoint (Mistral-7B).}
    \label{fig:raw_backbones}
\end{figure}
\subsection{Baselines}
We compare the \ourmodel{} family to other state-of-the-art $7$B and $13$B open-source models. Concretely, we include the following instruction and DPO finetuned variants of Mistral-7B \citep{jiang2023mistral}: \texttt{Mistral-7B-Instruct-v0.1}
, \texttt{Mistral-7B-Instruct-v0.2}
, as well as \texttt{zephyr-7b-beta}\footnote{\url{https://huggingface.co/HuggingFaceH4/zephyr-7b-beta}}. We also evaluate the Llama2 \citep{touvron2023llama} family, more specifically \texttt{Llama2-7b-Chat}
and \texttt{Llama2-13b-Chat}.
    
\subsection{Implementation Details}

\paragraph{Codebase} Our codebase relies on open-source frameworks \cite{shoeybi2019megatron,wolf2019huggingface,lhoest2021datasets} utilizing DeepSpeed (level 3) with Flash attention \cite{dao2022flashattention,dao2023flashattention}. It is built on PyTorch \cite{paszke2019pytorch}, and our models are available on the Huggingface hub.

\paragraph{Compute} Continuous pretraining utilizes $256$ MI250 AMD GPUs. For instruction fine-tuning, workload distribution occurs across 16 MI250. Evaluation procedures are seamlessly conducted on a single MI250.

\section{Results}

In this section, we discuss our main experimental findings and results.

\subsection{\legalbench{}}
\begin{figure}
    \centering
    \includegraphics[width=0.7\linewidth]{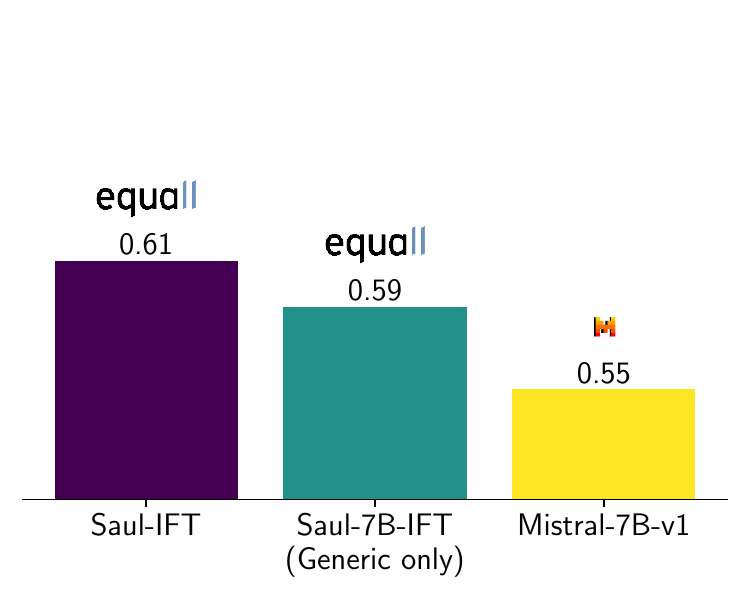}
    \caption{\textbf{Influence of the base model.} Starting the instruction finetuning from our base model \ourmodel{} brings noticeable improvements compared to the Mistral-7B. Indeed, even with a generic IFT mix (without legal), \ourmodel{} (Gen.) outperforms its Mistral-Instruct counterpart significantly. Adding legal instructions to the IFT mix further boosts the results.} 
    \label{fig:ablation_backbones}
\end{figure}

\Cref{fig:raw_backbones,fig:ablation_backbones} summarize our results on \legalbench{}. There are $3$ main takeaways, which we discuss below.

\paragraph{I. Legal continued pretraining brings significant improvements} We start by analyzing the impact of our proposed continued pretraining. As seen on \Cref{fig:raw_backbones}, \ourmodel{} is a strong standalone model. We speculate that its strong performance is largely due to the integration of instructions in the pre-training data, as mentioned in \autoref{sec:dataset_composition}. Nevertheless, we still note that even without a dedicated instruction fine-tuning stage, \ourmodel{} performs on par with \texttt{Llama2-7B-chat} ($0.38$ v.s. $0.39$). More importantly, \ourmodel{} serves as a strong base model for building IFT models with strong legal capabilities. When combined with Generic instruction finetuning, as seen on \Cref{fig:ablation_backbones}, it achieves a strong average of $0.59$, i.e. $4$ absolute points of improvement with respect to the best open-source instruct model \texttt{Mistral-7B-Instruct-v0.1}.

\begin{figure}
    \centering    \includegraphics[width=\linewidth]{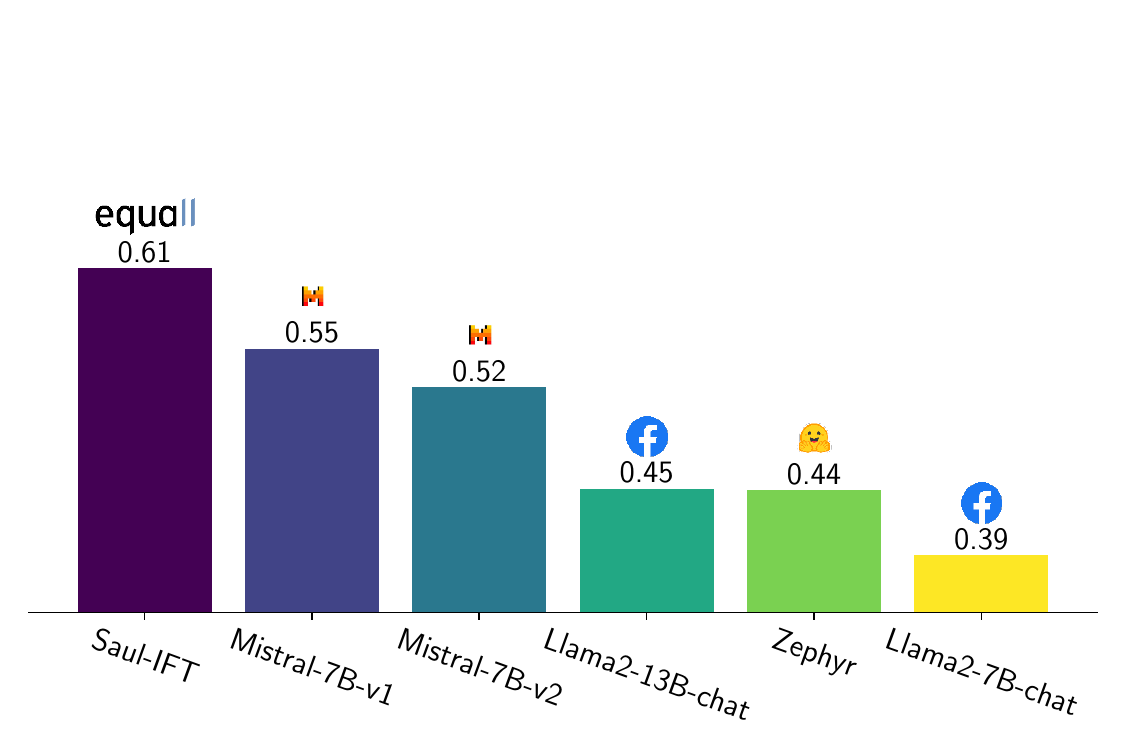}
    \caption{\textbf{Comparison of instruct models on \legalbench{}}. \ourmodelift{} establishes the state-of-the-art, outperforming the best Mistral-Instruct model by a significant 6 absolute points.}
    \label{fig:ift_comparison}
\end{figure}

\paragraph{II. Legal instruction finetuning further boosts the results} As seen on \Cref{fig:ift_generation}, finetuning \ourmodel{} on both general and legal instructions (\ourmodelift{}) establishes a new state-of-the-art on the \legalbench{} benchmark, with an average score of $0.61$, i.e. an $11$\% relative improvement compared to the best open-source instruct model (\Cref{fig:ift_comparison}. 
Finally, DPO-aligned models tend to underperform their instruction-tuned counterparts, which could be explained by the fact that generic alignment is not suited for out-of-distribution tasks, such as the ones present in \legalbench{}. Although beyond the scope of the present work, an interesting research direction would be to explore how legal-specific DPO can help.

\begin{figure}[!h]
        \centering
        \includegraphics[width=0.6\linewidth]{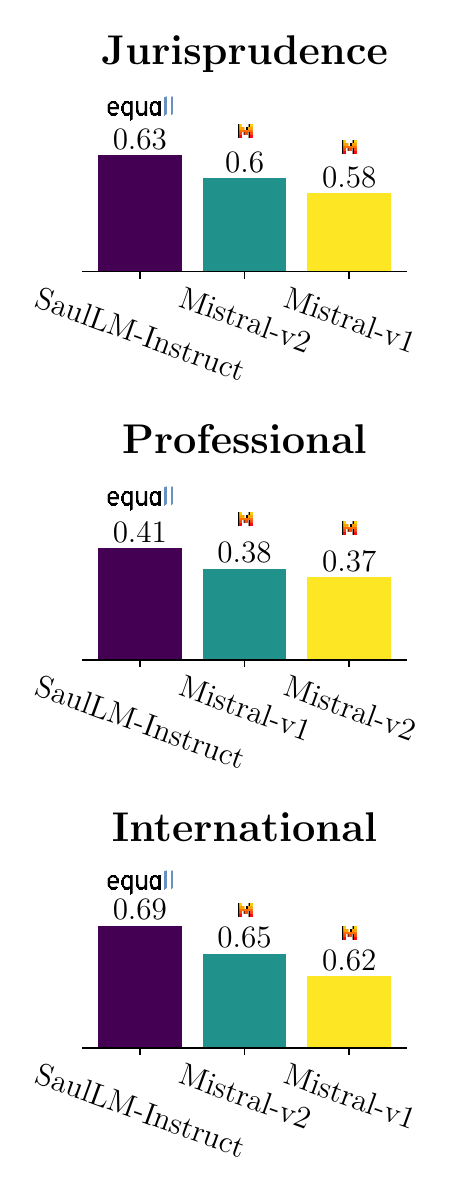}
        \caption{\textbf{Instruct models on \legalmmlu{}.} Echoing finding on \legalbench{}, \ourmodelift{} displays superior performance on all three tasks of \legalmmlu{}, with an average absolute improvement of 5 points with respect to \mistralinstructone. }
        \label{fig:mmlu_benchmark}
\end{figure}
\begin{figure}[!h]
    \centering
    \includegraphics[width=\linewidth]{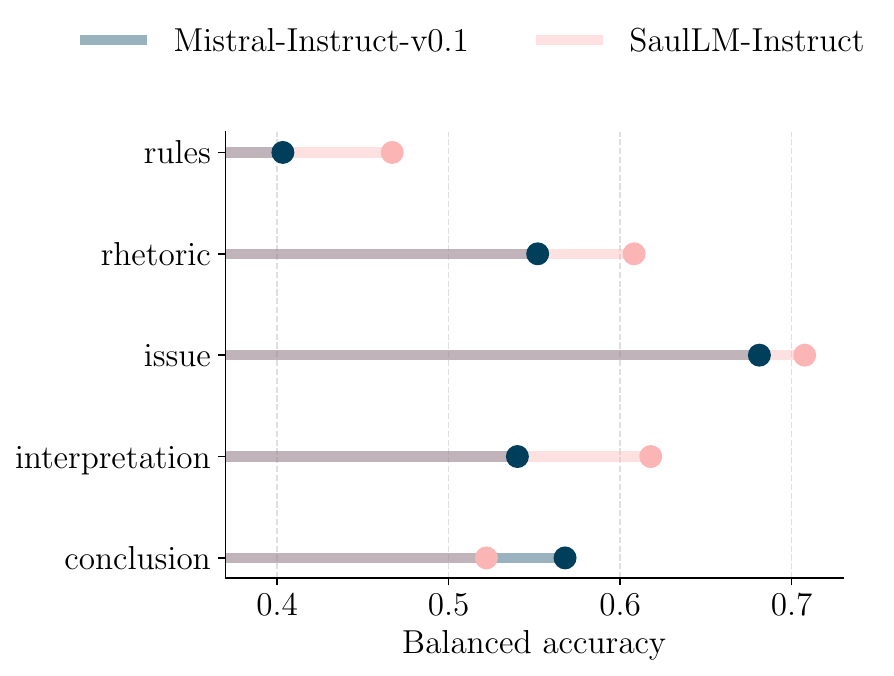}
    \caption{\textbf{Per-task performance breakdown.} \ourmodelift{} largely outperforms generic Instruct models on tasks that most require legal-specific knowledge, but is outperformed by Mistral-Instruct on the conclusion tasks, which necessitates more deductive reasoning.}
    \label{fig:spider_chat}
\end{figure}

\paragraph{III. There is still room for significant improvement.} Next, we follow the original LegalBench taxonomy \citep{guha2023legalbench} to gain a more granular understanding of \ourmodelift{}'s performance, by partitioning the tasks into $5$ core legal abilities: \textsc{Issue Spotting}, \textsc{Rule-Recall}, \textsc{Interpretation}, \textsc{Rhetoric Understanding}, and \textsc{Rule-Conclusion}. Results show an interesting trend (\Cref{fig:spider_chat}): \ourmodelift{} shows clear superior performance over the best non-legal competitor \texttt{Mistral-7B-Instruct-v0.1} on the four areas that require the most legal expertise, i.e. \textsc{Issue}, \textsc{Rule}, \textsc{Interpretation} and \textsc{Understanding}. On the other hand, it falls short of \texttt{Mistral-7B-Instruct-v0.1} on the \textsc{Conclusion} tasks, which interestingly require much more pure deductive reasoning than actual legal knowledge. We speculate that augmenting our pretraining and fine-tuning corpora with more deductive reasoning content, including but not limited to mathematics datasets could reduce the gap and fully unlock the potential of \ourmodelift{}.

\subsection{Results on \legalmmlu{}}

To confirm our observations on \legalbench{}, we analyze the results on \legalmmlu{} shown in \Cref{fig:mmlu_benchmark}. Again, \ourmodelift{} exhibits consistent superiority over non-legal instruction-tuned models, with a gap between $3$ and $4$ absolute points to the best 7B open-source competitor across the three tasks, providing additional evidence that \ourmodelift{} is as a strong foundation to build models tailored to legal workflows.
\begin{figure}[h]
        \centering
        \includegraphics[width=1\linewidth]{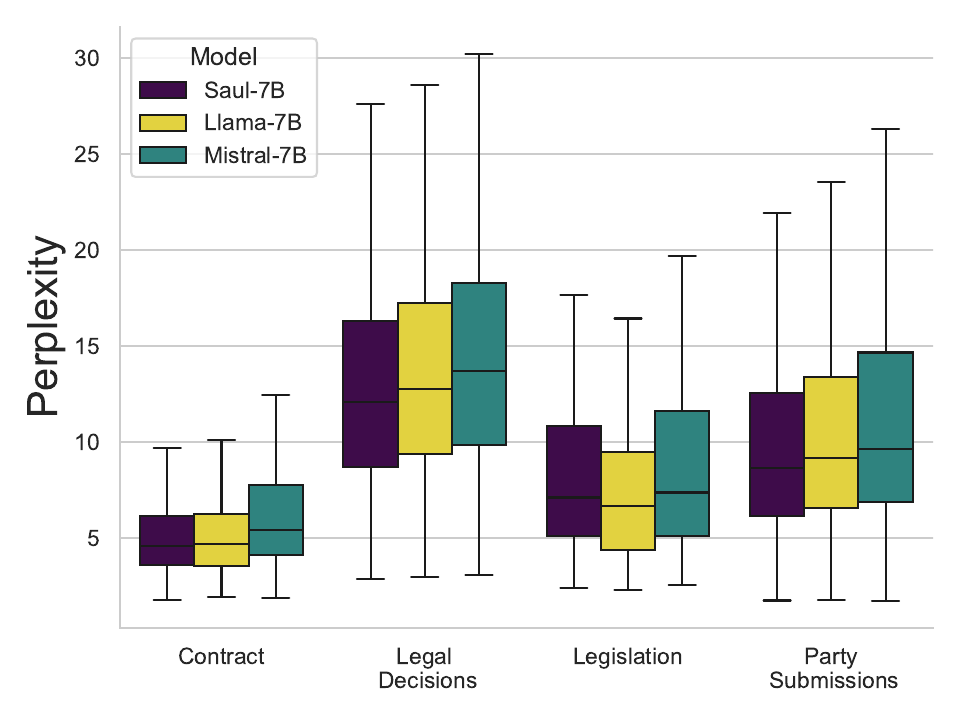}
        \caption{\textbf{Perplexity on legal documents for pretrained backbones.} \ourmodelift{} outperforms other pretrained backbones on most types of legal documents, but is outperformed by \texttt{Llama2-7b} on Legislation.
        \ourmodelift{} exhibits a median perplexity of $8.69$, having a reduction of $5.5$ percent compared to \texttt{Mistral-7B}, $9.20$, and $10.8$ percent compared to \texttt{Llama2-7B}, with a median perplexity of $9.74$.
        }
        \label{fig:perplexity_analysis}
\end{figure}

\subsection{Perplexity Analysis} 
To assess the adaptation of \ourmodel{} backbone to the legal domain, we present perplexity scores across four document types: contracts, legal decisions, legislation, and party submissions. Refer to \autoref{fig:perplexity_analysis} for the results. Our model, \ourmodel{}, consistently outperforms \texttt{Mistral-7B} across all categories, exhibiting lower average perplexity scores with reduced variance. Interestingly, \texttt{Llama2-7B} demonstrates lower perplexity specifically in legislation documents, suggesting a potentially higher proportion of legislative text in the pertaining corpora compared to \texttt{Mistral-7B}.

Overall, compared to \texttt{Mistral-7B}, our model shows a median perplexity reduction of 3 percent across legal corpora and 11 percent when compared to \texttt{Llama2-7B}.

\section{Conclusion \& Future Perspectives}

In this paper, we introduce \ourmodel{}, an open-source decoder model delivering state-of-the-art performance, compared to 7B models, within the legal domain. Our approach entails fine-tuning legal data alongside instruction fine-tuning on synthetic datasets. Additionally, we contribute by providing a cleaned version of LegalBench and introducing a new set of documents for perplexity measurement. We hope that our model, which is released under the MIT license, will contribute to the open-source ecosystem and the community.

\section*{Acknowledgments}
We thank GENCI for generously granting us access to their cutting-edge computing resources. Our model, \ourmodel{}, has been trained on ADASTRA, with initial experimentation conducted on Jeanzay. The utilization of HPC resources was made possible through the Jeanzay grants 101838, 103256, and 103298, as well as the Adastra grants C1615122, CAD14770, and CAD15031.

\bibliography{anthology,custom}

\appendix



\end{document}

%% file: Preamble/custom_packages.tex
\usepackage{hyperref}
\usepackage{cleveref}

\usepackage{todonotes}
\usepackage{xcolor}
\usepackage{xargs}
\usepackage{booktabs}
\usepackage{tablefootnote}

%% file: Preamble/custom_commands.tex
\definecolor{red}{RGB}{141,45,57}
\definecolor{dark}{RGB}{55,65,74}
\definecolor{blue}{RGB}{0,105,170}
\definecolor{gold}{RGB}{174,159,109}
\definecolor{gray}{RGB}{175,179,183}
\definecolor{darkgreen}{RGB}{50,110,30}

\newcommandx{\malik}[2][1=]{\vspace{0.2cm} \todo[linecolor=darkgreen,backgroundcolor=darkgreen!25,bordercolor=darkgreen, #1]{\textbf{Malik:} #2}}
\newcommandx{\pierre}[2][1=]{\vspace{0.2cm} \todo[linecolor=gold,backgroundcolor=gold!25,bordercolor=gold, #1]{\textbf{Pierre:} #2}}
\newcommandx{\telmo}[2][1=]{\vspace{0.2cm} \todo[linecolor=blue,backgroundcolor=blue!15,bordercolor=blue, #1]{\textbf{Telmo:} #2}}
\newcommandx{\melo}[2][1=]{\vspace{0.2cm} \todo[linecolor=blue,backgroundcolor=blue!15,bordercolor=blue, #1]{\textbf{Rui:} #2}}
\newcommandx{\dom}[2][1=]{\vspace{0.2cm} \todo[linecolor=purple,backgroundcolor=purple!15,bordercolor=purple, #1]{\textbf{Dominic:} #2}}
\newcommandx{\ourmodel}{\texttt{SaulLM-7B}}
\newcommandx{\ourmodelift}{\ourmodel\texttt{-Instruct}}
\newcommandx{\mistralinstructone}{\texttt{Mistral-7B-Instruct-v0.1}}
\newcommandx{\mistralinstructtwo}{\texttt{Mistral-7B-Instruct-v0.2}}
\newcommandx{\legalbench}{LegalBench-Instruct}
\newcommandx{\legalmmlu}{Legal-MMLU}